\documentclass[sigconf]{acmart}

\usepackage{multirow}
\usepackage{algorithm}
\usepackage{algorithmic}
\usepackage{tabularx}
\usepackage{amsmath}
\usepackage{amsthm}
\usepackage{amsfonts}
\usepackage{bm}

\AtBeginDocument{%
  \providecommand\BibTeX{{%
    \normalfont B\kern-0.5em{\scshape i\kern-0.25em b}\kern-0.8em\TeX}}}

\copyrightyear{2021}
\acmYear{2021}
\setcopyright{iw3c2w3}
\acmConference[WWW '21]{Proceedings of the Web Conference 2021}{April 19--23, 2021}{Ljubljana, Slovenia}
\acmBooktitle{Proceedings of the Web Conference 2021 (WWW '21), April 19--23, 2021, Ljubljana, Slovenia}
\acmPrice{}
\acmDOI{10.1145/3442381.3450064}
\acmISBN{978-1-4503-8312-7/21/04}

\begin{document}

\title{An Adversarial Transfer Network for Knowledge Representation Learning}

\author{Huijuan Wang}
\email{wanghj35@mail2.sysu.edu.cn}
\affiliation{
  \institution{School of Computer Science and Engineering, Sun Yat-sen University}
  \country{China}}

\author{Shuangyin Li}
\authornote{Rong Pan and Shuangyin Li are the corresponding authors.}
\email{shuangyinli@scnu.edu.cn}
\affiliation{
  \institution{Department of Computer Science, South China Normal University}
  \country{China}}

\author{Rong Pan}
\authornotemark[1]
\email{panr@sysu.edu.cn}
\affiliation{
  \institution{School of Computer Science and Engineering, Sun Yat-sen University}
  \country{China}}

\renewcommand{\shortauthors}{Wang, et al.}

\begin{abstract}
 Knowledge representation learning has received a lot of attention in the past few years. 
The success of existing methods heavily relies on the quality of knowledge graphs.
The entities with few triplets tend to be learned with less expressive power.
Fortunately, there are many knowledge graphs constructed from various sources, the representations of which could contain much information. 
We propose an adversarial embedding transfer network \textbf{ATransN}, which transfers knowledge from one or more teacher knowledge graphs to a target one through an aligned entity set without explicit data leakage.
Specifically, we add soft constraints on aligned entity pairs and neighbours to the existing knowledge representation learning methods. 
To handle the problem of possible distribution differences between teacher and target knowledge graphs, we introduce an adversarial adaption module. 
The discriminator of this module evaluates the degree of consistency between the embeddings of an aligned entity pair.
The consistency score is then used as the weights of soft constraints.
It is not necessary to acquire the relations and triplets in teacher knowledge graphs because we only utilize the entity representations.
Knowledge graph completion results show that ATransN achieves better performance against baselines without transfer on three datasets, CN3l, WK3l, and DWY100k.
The ablation study demonstrates that ATransN can bring steady and consistent improvement in different settings. 
The extension of combining other knowledge graph embedding algorithms and the extension with three teacher graphs display the promising generalization of the adversarial transfer network.
\end{abstract}




\maketitle

\section{Introduction}\label{sec:introduction}

Knowledge graphs are multi-relational directed graphs about facts, usually expressed in the form of triplets as ($h, r, t$), where $h,t$ are two entities and $r$ is the relation in between, e.g., (\textit{begin, Antonym, end}). 
Many applications ranging from recommendation~\cite{WangWX00C19} and question answering~\cite{kgqa20acl,LvGXTDGSJCH20} to machine reading comprehension~\cite{QiuZFLJLLZ19} benefit from such knowledge graphs. 
However, knowledge graphs often suffer from incompleteness.
For example, 75\% of persons in Freebase have no nationality~\cite{dong2014knowledge}.
Predicting such missing links is a crucial intrinsic task, called the knowledge graph completion task.

Representation learning for knowledge graph completion has recently received a lot of attention~\cite{DettmersMS018,ShangTHBHZ19,sun2019rotate}.
They focus on embedding entities and relations into vectors. 
Different models are designed based on triplets so that the learned embeddings could reflect the interactions among entities and relations.
Finally, missing relations can be predicted based on these embeddings.

Existing knowledge representation learning methods have shown superior performance on dense knowledge graphs. 
However, when the knowledge graph violates the triplets' density assumption, the performance will drop significantly. 
The embeddings of entities with insufficient triplets are rarely updated, and the expressiveness is limited~\cite{wang2016text}. 
Since the semantic features of a knowledge graph are limited, further progress requires external information. 
Some existing methods have introduced entity description~\cite{XieLJLS16,wang2016text}, but encoding text data will bring high computational costs. 
Another way to enrich the training set for low resource entities is to construct more correct triplets. 
Unfortunately, annotations are expensive and time-consuming in practice.

In order to improve the embedding quality without losing efficiency, we introduce an embedding transfer method, \textbf{A}dversarial \textbf{Trans}fer \textbf{N}etwork (\textbf{ATransN}). 
Like the applications of transfer learning in other fields~\cite{LiW0Z019,XuZNLWTZ20}, we need a set of pre-trained knowledge graph embeddings in the teacher domain.
The teacher knowledge graph contains more information and has a set of entities aligned with the target knowledge graph.
ATransN is designed for shallow knowledge graph embedding models such as TransE~\cite{bordes2013translating}, whose objective function only depends on triplets.
Considering data security, ATransN only acquires the entity embeddings so that we cannot recover many facts of the teacher knowledge graph when relation information is unknown.

\begin{figure}[tbp]
\centering
\includegraphics[width=0.95\columnwidth]{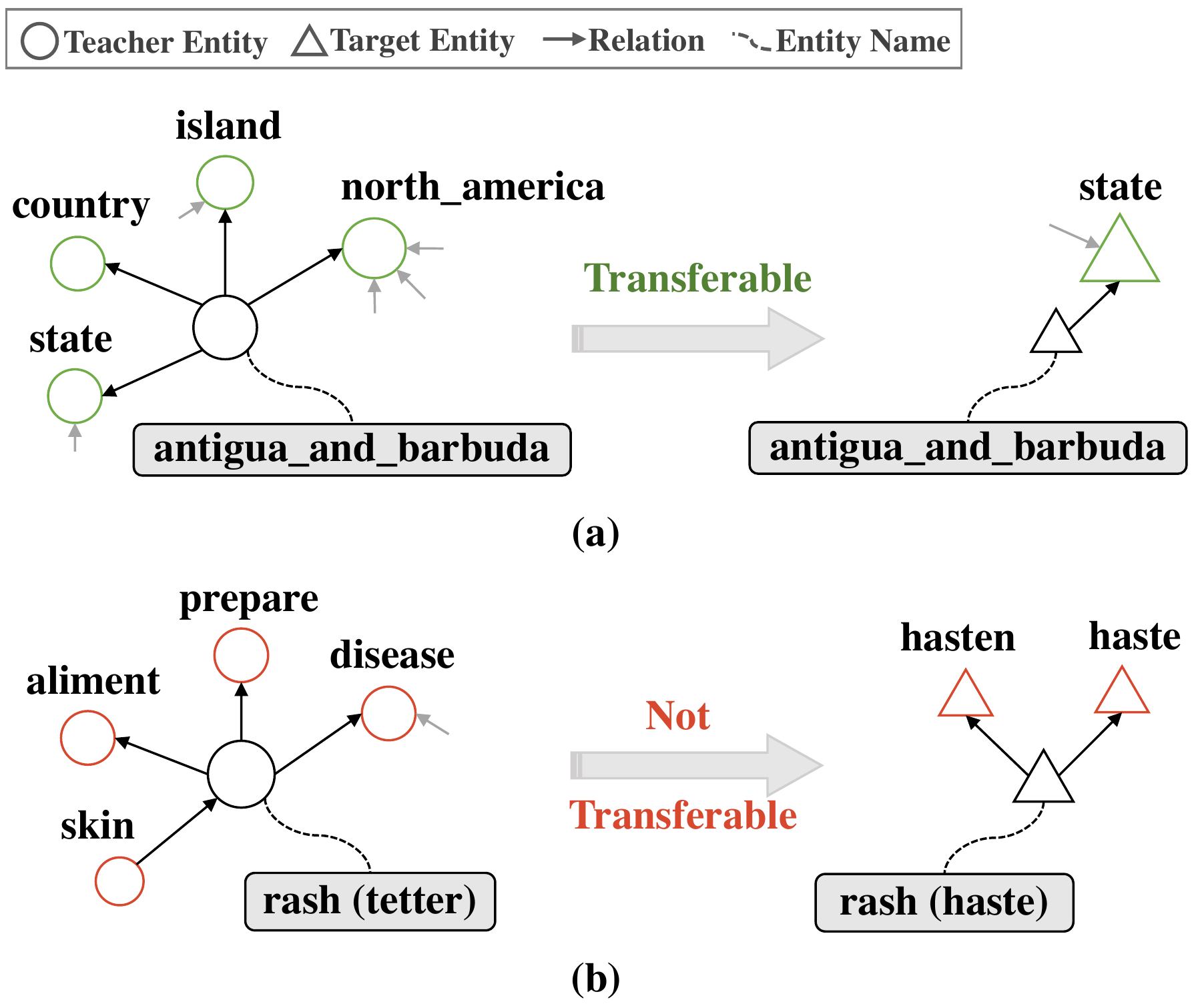}
\caption{Examples of two transfer situations. 
(a) Useful case. 
Neighbours of \textit{antigua\_and\_barbuda} in both knowledge graphs are related to ``country", where the relevant neighbours are denoted as green.
(b) Useless case. 
Most neighbours of \textit{rash} in the teacher knowledge graph are about ``tetter", while \textit{rash} in the target knowledge graph means ``haste". 
The irrelevant neighbours are denoted as red.}
\label{fig:example}
\Description[Two opposite cases in the transfer process]{(a) shows neighbour entities around antigua\_and\_barbuda (entity), including ``country'', ``state'', ``north_amarica'' in the teacher knowledge graph and ``state'' in the target knowledge graph; (b) shows unrelated neighbours of the entity rash in both the teacher and the target knowledge graphs.}
\end{figure}

Figure~\ref{fig:example} shows two different cases of aligned entities during the transfer. 
The neighbour entities of \textit{antigua\_and\_barbuda} in Figure~\ref{fig:example} (a) are both related to the concept of ``country", such as the common neighbour ``state'' in both knowledge graphs.
If the overlapped neighbours have related semantic meanings, then the teacher embedding is helpful to the target knowledge graph.
Based on this intuition, we expect the entity embeddings in the target domain to be as close as possible to the aligned entity embeddings in the teacher domain, besides the original objective on triplets in the target knowledge graph. 
Such an assumption can be implemented as two constraints.
First, define a distance function between the aligned entity pair and make the distances as close as possible. 
Second, assume that a triplet in the target knowledge graph still holds after replacing one entity with the aligned teacher entity and minimize the new transferred triplet scores. 
The former way transfers features in the teacher embeddings through updating the aligned entity embeddings, while the latter acts more on the neighbour entities and relations.
Besides, the latter is more general as it is difficult to define a proper distance function in some cases.

Meanwhile, a disjoint neighbour set in the teacher domain will contribute irrelevant features to the aligned entity embeddings in the target domain.
For example, in Figure~\ref{fig:example} (b), the meaning ``tetter" of rash in the teacher knowledge graph is different from the meaning ``haste" in the target knowledge graph. 
It is inevitable because the teacher and the target knowledge graphs are related but not the same under our assumption. 
This distribution difference brings uncertainty during the embedding transfer. 
It has been shown that brute-force transfer may hurt the performance of learning in the target domain~\cite{PanY10}. 
To avoid such negative transfer, we introduce an adversarial adaptation module to filter out irrelevant features in the transferred embeddings. 
Specifically, a discriminator tries to distinguish the transferred embeddings' distribution and the target embeddings' distribution, and evaluates consistency score between the transferred teacher embedding and the target embedding.
The consistency score is used as the weight of the two constraints above.
A generator generates noisy transferred embeddings from {conditional} signals to improve the evaluation performance.

The contributions of this paper can be highlighted as follows. 
First, we extend the knowledge graph embedding methods with adversarial transfer learning under the ATransN framework.
Second, we demonstrate that ATransN successfully makes good use of teacher knowledge graph embeddings to improve the knowledge graph completion performance on three different knowledge graphs. 
Third, we conduct exhaustive ablation studies to analyze each module's importance in ATransN, finding both soft constraints and the adversarial adaptation module have positive effects on the knowledge graph completion task. 
Last, we show that ATransN is a general and promising framework for knowledge graph completion by extending to other knowledge graph embedding algorithms or multiple knowledge graphs as teachers at the same time.
Code and data are released in \url{https://github.com/LemonNoel/ATransN}.

\section{Related Work}\label{sec:relatedwork}

Previous triplet-based knowledge representation learning methods could be roughly divided into shallow models and deep models.
Given a knowledge graph, shallow models define score functions for triplets according to different assumptions on the graph structure.
For example, translation-based models~\cite{bordes2013translating,wang2014knowledge,lin2015learning} assume that the relationship between two entities corresponds to a translation between the two entity embeddings.
Bilinear models~\cite{yang2014embedding,trouillon2016complex} model entities and relations in triplets by matching semantics in the vector space.
Some work extends real-valued vectors to complex-valued vectors~\cite{sun2019rotate} and hypercomplex-valued vectors~\cite{zhang2019quaternion} so that interactions can be modeled compactly, e.g., as rotation.
Shallow models are always simple to implement and could have a competitive performance for specific knowledge graphs. 
Deep models tend to have better modeling abilities in theory, but require higher complexities of time and space because of the neural network besides entity and relation embeddings. 
~\citeauthor{DettmersMS018}~\shortcite{DettmersMS018} use a convolutional neural network to extract features from a head-relation pair and predict tail entities.
RGCN~\cite{schlichtkrull2017modeling} and CompGCN~\cite{VashishthSNT20} encode the neighbour structure around entities with graph neural networks. 
We exclude deep models in our framework as it is challenging to analyze the expressiveness of learned embeddings.

Although embedding methods above have exhibited superior performance in the knowledge completion task, they face poor performance when graph data is sparse. 
Under the circumstances, embeddings of long-tail entities and relations are rarely updated so that these triplet-based methods' performance may decrease.
Hence, additional information beyond triplets is introduced as supplementary, including entity types~\cite{XieLS16}, textual descriptions~\cite{XieLJLS16} and images~\cite{XieLLS17}. 
However, they only work on knowledge graphs with corresponding annotations, and encoding supplemental data is time-consuming.
Some information obtained in unsupervised ways is also useful, including context words~\cite{WangZFC14}, relation paths~\cite{GuML15}, and even logical rules~\cite{WangWG15}.
Beyond a single knowledge graph, some work~\cite{LiangLZS19} has tried to transfer relations for clustering.
In this paper, we use semantic features hidden in pre-trained embeddings from auxiliary teacher knowledge graphs to improve triplet-based embedding methods. 
This framework only requires an aligned entity set between two knowledge graphs or multiple aligned sets for multiple teacher knowledge graphs, which can be constructed using string matching or interlinks between knowledge graphs.

Transfer learning can be implemented in different ways.
Instance-based transfer learning reuses data of source domain in the target learning~\cite{DaiYXY07}, while the feature-based aims to transfer knowledge across domains through feature encoding~\cite{BlitzerMP06}. 
For knowledge representation learning, it is expensive to retrain a large-scale auxiliary knowledge graph when reusing data.
Besides, recollecting source triplets is sometimes impossible considering data security. 
Therefore, we focus on transferring the learned embeddings' features of the auxiliary knowledge graph. 
The adversarial module used to improve the transfer process, has also shown excellent efficiency in other knowledge graph tasks, such as negative sampling~\cite{WangLP18} and knowledge graph alignment~\cite{qu2019weakly}.

\section{Notations}

The framework involves two or more different knowledge graphs.
A target knowledge graph is denoted as $\mathcal{G} = (\mathcal{E}, \mathcal{R}, \mathcal{S})$, where $\mathcal{E}$ is the entity set, $\mathcal{R}$ is the relation set, and $\mathcal{S}$ is the triplet set.
Each triplet $(h, r, t) \in \mathcal{S}$ represents a relation $r$ between a head entity $h$ and a tail entity $t$.
Without loss of generality, we introduce the situation of one teacher.
A teacher knowledge graph is another directed graph $\mathcal{G}_t$. 
Commonly, the entity set of the teacher knowledge graph is different from that of the target knowledge graph, let alone relations. 
It is not trivial to acquire teacher knowledge graph data due to data security. 
However, the pre-trained entity representations can be available because we cannot recover many facts without relation types and relation representations.
To utilize these representations in the teacher knowledge graph, we also need entity alignment information. An aligned entity pair set is denoted as $\mathcal{C} = \{ (e_t, e_s) \}$ where $e_t$ comes from the teacher knowledge graph, $e_s$ comes from the target knowledge graph, and both of them refer to the same entity.
Corresponding embeddings are denoted in bold.

\begin{figure}[tbp]
  \centering
  \includegraphics[width=\linewidth]{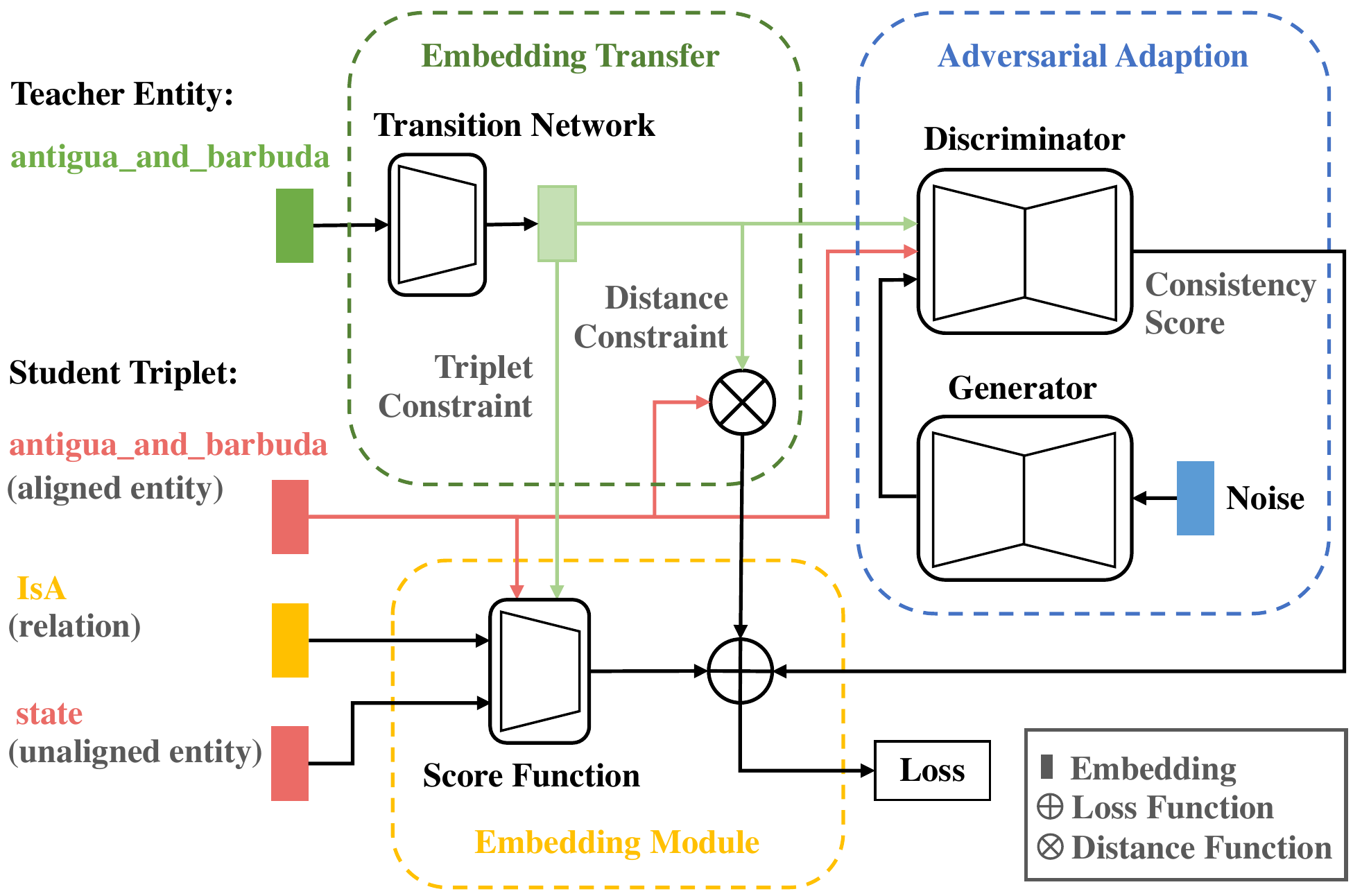}
  \caption{Framework Overview. ATransN consists of three modules, including embedding module, embedding transfer module and adversarial adaption module.}
  \label{fig:arc}
  \Description[Framework of ATransN.]{}
\end{figure}

\section{Adversarial Transfer Network}

Given entity embeddings of a teacher knowledge graph and a target knowledge graph, the goal of the Adversarial Transfer Network is to learn the entity and relation embeddings of the target knowledge graph under soft constraints of the teacher representations.

As illustrated in Figure~\ref{fig:arc}, the framework consists of three modules: (1) an \textbf{embedding module} aiming to learn representations from triplets in the target knowledge graph; (2) an \textbf{embedding transfer module} aligning teacher entities and target entities through a transition network, a distance constraint, and a transferred triplet constraint; (3) an \textbf{adversarial adaptation module} evaluating the degree of consistency between an aligned entity pair to make constraints soft.

\subsection{Embedding Module}

Our framework extends shallow knowledge representation learning models with transfer learning and adversarial learning. 
As mentioned in Section~\ref{sec:relatedwork}, shallow models always define score functions based on triplets. 
The original translation-based model TransE~\cite{bordes2013translating} defines a score function as Eq.~\eqref{eq:transe}.
It follows a simple assumption that the addition of a head entity embedding and a relation embedding equals the tail entity embedding.
This method still outperforms many latter shallow models with proper hyper-parameters.
\begin{align}
    f_r (\bm{h}, \bm{t}) = {\parallel \bm{h} + \bm{r} - \bm{t} \parallel }. \label{eq:transe}
\end{align}

Except for score functions, there are many skills proposed to improve knowledge representation learning. 
Negative sampling has been proved quite efficient in previous studies~\cite{bordes2013translating,trouillon2016complex}.
In this paper, we follow the ``unif" strategy~\cite{wang2014knowledge}:
given a valid triplet ($h, r, t$), negative triplets ($h',r,t'$) are drawn by replacing the head or tail entity with an entity randomly sampled from $\mathcal{E}$ with an equal probability. 
The negative triplet set is denoted as $\mathcal{S}'$ and the sampled distribution is $p_{s'|(h,r,t)}$.
We use the training objective from RotatE~\cite{sun2019rotate} for effectively optimizing distance-based models based on negative sampling loss~\cite{MikolovSCCD13}:
\begin{align}\label{embeddingmodel}
    \mathcal{L}_e & = \mathbb{E}_{(h,r,t) \sim p_s}[-\log(\sigma(\gamma - f_r({\bm{h}, \bm{t}})) \nonumber \\
    & \quad  - \mathbb{E}_{(h',r,t') \sim p_{s'|(h,r,t)}}[\log(\sigma(f_r(\bm{h}', \bm{t}') - \gamma))]],
\end{align}
where $\gamma$ is the fixed margin, $\sigma$ is the Sigmoid function, $p_{s}$ is the distribution of $\mathcal{S}$.

\subsection{Embedding Transfer Module}

Shallow knowledge representation learning models always assume that knowledge graphs are dense enough. 
However, there are long-tail entities and relations and these corresponding representations are rarely updated in practice.
Thus, the quality of learned representations declines as the number of triplets reduces.
We introduce an embedding transfer module, which aims to transfer features learned in the teacher knowledge graph to the target one through the aligned entity set $\mathcal{C}$. Aligned entities and long-tail relations in the target knowledge graph could benefit from the transferred features.
There are two ways to implement the transfer process.

First, we could construct a \textbf{distance constraint} based on the aligned entity set $\mathcal{C}$. 
Embeddings of aligned entity pair $(e_t, e_s)$ possibly have different dimensions $m$ and $n$. 
To handle this problem, teacher entity embeddings are fed to a transition network denoted as $W: \mathbb{R}^m \rightarrow \mathbb{R}^n$. 
In this case, teacher entity embeddings are projected into the target space by the transition network. 
Then the projected teacher embeddings are taken as soft targets of the corresponding target entities. 
We formulate such a constraint as a distance function $f_d(\mathbf{\mathcal{C}})$ defined with the projected teacher embeddings and the target embeddings of aligned entity pairs as follows:
\begin{align}
    f_d (\mathbf{\mathcal{C}}) = \sum_{(e_t, e_s) \in \mathcal{C}}{ \phi(\bm{e_t}, \bm{e_s})}, \label{eq:emb_reg}
\end{align}
where $\phi$ is a distance function to evaluate the distance between embeddings. 
We assume that the aligned entity embeddings in different knowledge graphs tend to be close in the target space when they are consistent with each other.
We choose the Cosine distance as $\phi$, and it is defiend as $\phi(\bm{e_t}, \bm{e_s}) = 1 - \cos(W (\bm{e_t}), \bm{e_s})$.

Second, we can utilize relations and neighbour entities in the target knowledge graph. 
We assume that a target triplet assumption also holds after replacing one entity embedding with the corresponding projected teacher entity embedding.
More specifically, given a triplet $(h, r, t)$ in the target knowledge graph $\mathcal{G}$, we assume that the aligned teacher entities $\{e_t\}$ should also comprise valid triplets \{($e_t, r, t$)\} if $e_t$ aligns with $h$ or \{($h, r, e_t$)\} if $e_t$ aligns with $t$. We denote these valid triplets as transferred triplets.
Similar to the first constraint, we project teacher embeddings to the target space through the transition network $W$. 
The goal is to make the transferred triplets fit the score function in the embedding module.
Hence, we formulate the \textbf{triplet constraint} as Eq.~\eqref{eq:triplet_reg}, which is to minimize scores of the transferred triplets: 
\begin{align}
    f_n(\mathcal{S}, \mathcal{C}) & = \mathbb{E}_{(h, r, t) \sim p_s}[\mathbb{E}_{(e_t, h) \sim p_c}[- \log(\sigma(\gamma - f_r(W (\bm{e_t}), \bm{t})))] \nonumber \\ 
    & \qquad \quad + \mathbb{E}_{(e_t, t) \sim p_c}[- \log(\sigma(\gamma - f_r(\bm{h}, W (\bm{e_t}))))]], \label{eq:triplet_reg}
\end{align}
where $\gamma$ is the margin used in the embedding module, $p_{c}$ is the distribution of $\mathcal{C}$.

Finally, we define the training objective of the embedding module and the embedding transfer module as follows:
\begin{align}
    \mathcal{L} = \mathcal{L}_e + \alpha f_d(\mathcal{C}) + \beta f_n(\mathcal{S}, \mathcal{C}), 
    \label{eq:obj}
\end{align}
where $\alpha$ and $\beta$ are hyper-parameters to control the weights of the transferred embedding distance constraint and the transferred triplet constraint.

\subsection{Adversarial Adaptation Module}

The transfer process is the key component of ATransN. However, semantic meanings of aligned entity pairs are not always consistent as illustrated in Figure~\ref{fig:example}. The more similar the neighbours are, the stronger supervision should the constraints provide. Thus, it is better to assign a dynamic weight to constraints during the embedding transfer according to the degree of consistency between an aligned entity pair.

\begin{table*}[!htbp]
  \centering
  \caption{Statistics of datasets. \textit{Alignment ratio} means the ratio of the number of aligned entities to all entities in a domain.}
  \label{tab:stts}
  \begin{tabular}{ccc|ccccc}
    \toprule
     Data & Teacher & Target & \#Entities & \#Relations & \#Triplets & \#Aligned entities & Alignment ratio (\%) \\
    \midrule 
    \multirow{2}{*}{CN3l} & {ConceptNet (EN)} &  & 4,316 & 43 & 32,528 & 4,043 & 93.67 \\
     &  & ConceptNet (DE) & 4,302 & 7 & 12,780 & 3,908 & 90.84 \\
    \hline
    \multirow{2}{*}{WK3l-15k} & Wikipedia (EN) &  & 15,169 & 2,228 & 203,502 & 2,496 & 16.45 \\
     &  & Wikipedia (FR) & 15,393 & 2,422 & 170,605 & 2,458 & 15.97 \\
    \hline
    \multirow{4}{*}{DWY100k} & DBpedia (WD) & & 100,000 & 330 & 463,294 & 100,000 & 100.00 \\
     & & Wikidata & 100,000 & 220 & 448,774 & 100,000 & 100.00  \\
    \cline{2-8}
     & DBpedia (YG) & & 100,000 & 302 & 428,952 & 100,000 & 100.00 \\
     & & YAGO & 100,000 & 31 & 502,563 & 100,000 & 100.00  \\
    \bottomrule
  \end{tabular}
\end{table*}

Motivated by this idea, we add an adversarial adaption module to evaluate the degree of consistency between an aligned entity pair, where the discriminator gives the consistency score, and the generator generates noisy transferred embeddings to improve the discriminator~\cite{GoodfellowPMXWOCB14}. 
A naive generator generates fake examples $z$ from a noise prior distribution $p_z$ and hopes $G(z)$ could become a good estimator of the target entity distribution $p_e$. 
It simply initializes the distribution $p_z$ as a uniform distribution or normal distribution if there is no prior knowledge.
However, the entity embedding space is unlimited so that such a method always fails.
Inspired by the Conditional GAN~\cite{MirzaO14}, we assume the prior noise distribution is a standard uniform distribution $\mathcal{U}(-1, 1)$, and use a linear layer with input $e$ and the sampled uniform noise to shape the conditional distribution $p_{z|e}$:
\begin{align}
    \mathcal{L}_g = \mathbb{E}_{e \sim p_e, \bm{z} \sim p_{z}}[- \log(D(\bm{e}, G(\bm{e}, \bm{z})))], \label{eq:generator}
\end{align}
where $z$ is a sampled from the standard uniform distribution $p_{z}$, $G(\bm{e}, \bm{z})$ is a conditional signal following $p_{z|e}$, $D$ is a discriminator is used to measure the embedding consistency of two aligned entities, $D(\bm{e}, G(\bm{e}, \bm{z}))$ is a score to the degree of consistency. 
A new issue is that the embedding space may be not closed so that $p_{z|e}$ arises the instability problem. Hence, we also add the cosine distance constraint between the $\bm{e}$ and $\bm{z}$.
The binary cross-entropy is used to train the discriminator as Eq.~\eqref{eq:discriminator}.
Once we get the output of the discriminator, we can use the score as the weight of the embedding transfer module.
A larger score means two entities are more consistent so that features in the teacher knowledge graph are more useful.
\begin{align}
    \mathcal{L}_d &= -\mathbb{E}_{(e_t, e_s) \sim p_c}[\log(D(\bm{e}_s, W(\bm{e_t})))]  \nonumber \\
    & \qquad - \mathbb{E}_{e \sim p_e, \bm{z} \sim p_{z}}[\log(1-D(\bm{e}, G(\bm{e}, \bm{z})))]. \label{eq:discriminator}
\end{align}

Therefore, Eq.~\eqref{eq:emb_reg} and Eq.~\eqref{eq:triplet_reg} can further benefit from the discriminator.
The discriminator output can guide whether the aligned embeddings could help target knowledge graph representation learning.
We add the output as the weights for Eq.~\eqref{eq:emb_reg} and Eq.~\eqref{eq:triplet_reg}. Eq.~\eqref{eq:adj_embed_reg} is the adjusted distance function, and Eq.~\eqref{eq:triplet_reg} can also be adjusted in the similar way. 
\begin{align}
    f_d (\mathbf{\mathcal{C}}) = \sum_{(e_s, e_t) \in \mathcal{C}}{D(\bm{e_s}, W(\bm{e_t})) \cdot \phi(\bm{e_s}, \bm{e_t})}. \label{eq:adj_embed_reg}
\end{align}

\section{Experiments}\label{sec:experiments}

\subsection{Knowledge Graph Completion}

Knowledge graph completion aims to predict the missing entity in a triplet, namely to predict $h$ given ($r, t$) or $t$ given ($h, r$) as defined in~\cite{bordes2013translating}. 
It reflects the expressiveness of the embeddings learned by a model. 
For each positive test triplet ($h, r, t$), we replace the head (or tail) entity with all entities in $\mathcal{E}$ to construct corrupted triplets. 
Then we compute the triplet scores of the ground-truth triplet and its corresponding corrupted triplets. 
Scores are further sorted in ascending order so that we can obtain metrics based on ranking.
We report the results on four metrics, including Mean Rank (MR), Mean Reciprocal Rank (MRR), Hits@3, and Hits@10, where Hits@$K$ denotes the proportion of correct entities ranked in top $K$.
A lower Mean Rank, a higher Mean Reciprocal Rank, or a higher Hits@$K$ usually means better performance. 
Since a corrupted triplet might also exist in the target knowledge graph, these metrics will be adversely affected. 
To avoid underestimating the performance of models, we remove all the corrupted triplets that already exist in the target knowledge graph (including training, validation, and test parts) and take the filtered rank of the positive triplet, which denoted as the ``filter" setting~\cite{DettmersMS018}.

\begin{table}[htbp]
    \small
    \centering
    \caption{Score functions of baselines.}
    \label{tab:baselines}
    \begin{tabular}{c|c|c}
        \toprule
         Method & Score Function & Remarks \\
         \hline
         TransE & $f_r (\bm{h}, \bm{t}) = {\parallel \bm{h} + \bm{r} - \bm{t} \parallel} $ & $\bm{h}, \bm{r}, \bm{t} \in \mathbb{R}^{d}, {\parallel \bm{h} \parallel} = 1, {\parallel \bm{t} \parallel} = 1$ \\
         DistMult & $-\bm{h}^\top \text{diag}(\bm{r}) \bm{t}$ & $\bm{h}, \bm{r}, \bm{t} \in \mathbb{R}^{d}$ \\ 
         ComplEx & $-\text{Re}(\bm{h}^\top \text{diag}(\bm{r}) \overline{\bm{t}})$ & $\bm{h}, \bm{r}, \bm{t} \in \mathbb{C}^{d}$ \\ 
         RotatE & $\parallel \bm{h} \circ \bm{r} - \bm{t} \parallel$ & $\bm{h}, \bm{r}, \bm{t} \in \mathbb{C}^{d}, \forall i \ |r_i|=1$ \\
         \bottomrule
    \end{tabular}
\end{table}

\begin{algorithm}[t]
\caption{Training process of ATransN.}
\label{algorithm1}
{\small
\begin{algorithmic} 
\REQUIRE Target training data $\mathcal{T}=(h,r,t)$,
teacher entity embedding set $\mathcal{E}_s=\{\bm{e}_s\}$, aligned entity set $\mathcal{C}=\{(e_s, e_t)\}$, overall training steps $T_l$, training steps of the generator $T_g$, training steps of the discriminator $T_d$, negative sampling size $k$, minibatch size for the embedding module $N_l$, minibatch size for the adversarial adaptation module $N_a$.
\STATE {\textbf{Training:}}
\STATE {Initialize target entity and relation embeddings with uniform distributions}
\FOR{$i \gets 1 \ \TO \ T_{l}$}

\FOR{$j \gets 1 \ \TO \ T_{d}$}
\STATE{Sample $N_a$ aligned pairs \{$(e_s, e_t)^{(1)}, \cdots, (e_s, e_t)^{(N_a)}$\} from $\mathcal{C}$}
\STATE{Sample 1 noisy sample $\bm{z}^{(i)}|e_s^{(i)}$ from conditional distribution $p_{z|e_s}$ for each pair $(e_s, e_t)^{(i)}$ respectively.}
\STATE{Update the discriminator by descending its stochastic gradient of Eq.~\eqref{eq:discriminator}}
\ENDFOR

\FOR{$j \gets 1 \ \TO \ T_{g}$}
\STATE{Sample $N_a$ aligned pairs \{$(e_s, e_t)^{(1)}, \cdots, (e_s, e_t)^{(N_a)}$\} from $\mathcal{C}$}
\STATE{Sample 1 noisy sample $\bm{z}^{(i)}|e_s^{(i)}$ from conditional distribution $p_{z|e_s}$ for each pair $(e_s, e_t)^{(i)}$ respectively.}
\STATE{Update the generator by ascending its stochastic gradient of Eq.~\eqref{eq:generator}) with the distance constraint}
\ENDFOR

\STATE{Sample $N_l$ triplets $\{(h, r, t)^{(1)}, \cdots, (h, r, t)^{(N_l)}\}$ from $\mathcal{T}$}
\STATE{Sample $k$ negative samples by replacing $h$ or $t$ for each triplet $(e_s, e_t)^{(i)}$ respectively}
\STATE{Construct transferred triplets $\{(e_s, r, t)^{(1)}, \cdots\}$ by replacing $h$ with $e_s$ if $(e_s, h) \in \mathcal{C}$ and $\{(h, r, e_s)^{(1)}, \cdots\}$ by replacing $t$ with $e_s$ if $(e_s, t) \in \mathcal{C}$}
\STATE{Sample $N_a$ aligned pairs \{$(e_s, e_t)^{(1)}, \cdots, (e_s, e_t)^{(N_a)}$\} from $\mathcal{C}$}
\STATE{Update the embedding module by descending its stochastic gradient of Eq.~\eqref{embeddingmodel}}
\ENDFOR
\ENSURE The entity and relation embeddings of the embedding module.
\end{algorithmic}
}
\end{algorithm}

\subsection{Datasets}

We conduct experiments on three benchmarking datasets, {CN3l (EN-DE), WK3l-15k (EN-FR)}~\cite{Chen_2017}\footnote{\url{https://github.com/muhaochen/MTransE}}, and DWY100k~\cite{BootEA}\footnote{\url{https://github.com/nju-websoft/BootEA}}, all of which are originally constructed for the entity alignment task.
Table~\ref{tab:stts} summarizes data statistics.
In this paper, each dataset is randomly split into three parts where 60\% triplets as the training data, 20\% triplets as the validation data, and 20\% triplets as the test data.

\begin{itemize}
    \item \textbf{CN3l (EN-DE)} is constructed from ConceptNet~\cite{speer2013conceptnet}, containing an English knowledge graph and a German knowledge graph.
    The aligned entities are linked according to the relation \textit{TranslationOf} in ConceptNet.
    As there are fewer German triplets per entity, we take the English knowledge graph as the teacher and learn embeddings with methods such as TransE. 
    Then, we use ATransN to learn entities and relation embeddings based on the German triplets as well as the entity embeddings of the English knowledge graph.
    \item \textbf{WK3l-15k (EN-FR)} is created from Wikipedia, including an English knowledge graph and a French knowledge graph. 
    The aligned entity set is constructed by verifying the inter-lingual links.
    As the English knowledge graph has more triplets than the French, we also use the English knowledge graph as the teacher knowledge graph and the French knowledge graph as the target. 
    \item DWY100k contains two large-scale datasets constructed from three data sources, DBpedia, Wikidata and YAGO. The two datasets are denoted by \textbf{DBP-WD} and \textbf{DBP-YG}, where all entities are 100\% aligned. Although the YAGO knowledge graph has more triplets, we both take the DBpedia knowledge graph as the teacher knowledge graph here.
\end{itemize}

\subsection{Baselines}

We compare ATransN with several competitive baselines mentioned in Section~\ref{sec:relatedwork}, including TransE~\cite{bordes2013translating}, DistMult~\cite{yang2014embedding}, ComplEx~\cite{trouillon2016complex}, and RotatE~\cite{sun2019rotate}. 
Score functions of these models are listed in Table~\ref{tab:baselines}.
We implement all models under PyTorch\footnote{\url{https://pytorch.org}} framework.

\begin{table*}[!ht]
  \centering
  \caption{Model performance of different embedding models on CN3l and WK3l-15k.}
  \label{tab:cn3l_wk3l}
  \begin{tabular}{c|cccc|cccc}
    \toprule
    & \multicolumn{4}{c|}{CN3l (EN-DE)} & \multicolumn{4}{c}{WK3l-15k (EN-FR)} \\
    Model & MR & MRR & Hits@3 (\%) & Hits@10 (\%) & MR & MRR & Hits@3 (\%) & Hits@10 (\%) \\
    \midrule
    TransE & 910 & 0.162 & 26.29 & 35.76 & 443 & 0.419 & 45.39 & 58.83 \\
    DistMult & 1,333 & 0.179 & 19.23 & 23.14 & 1,202 & 0.331 & 35.97 & 47.29 \\
    ComplEx & 1,481 & 0.133 & 13.99 & 17.45 & 2,079 & 0.301 & 32.28 & 40.27\\
    RotatE & 822 & \bf 0.229 & 26.64 & 35.11 & 483 & 0.392 & 42.28 & 56.03 \\
    \midrule
    ATransN & \bf 446 & 0.205 & \bf 33.76 & \bf 46.48 & \bf 403 & \bf 0.422 & \bf 45.75 & \bf 59.30 \\
    \bottomrule
  \end{tabular}
\end{table*}

\begin{table*}[!ht]
  \centering
  \caption{Model performance of different embedding models on DWY100k.}
  \label{tab:dwy100k}
  \begin{tabular}{c|cccc|cccc}
    \toprule
     & \multicolumn{4}{c|}{DWY100k (DBP-YG)} & \multicolumn{4}{c}{DWY100k (DBP-WD)} \\
    Model & MR & MRR & Hits@3 (\%) & Hits@10 (\%) & MR & MRR & Hits@3 (\%) & Hits@10 (\%)\\
    \midrule
    TransE & 2,778 & 0.220 & 25.65 & 38.49 & 3,148 & 0.237 & 33.03 & 44.16 \\
    DistMult & 11,221 & 0.209 & 22.38 & 30.64 & 16,276 & 0.164 & 19.73 & 27.99 \\
    ComplEx & 19,932 & 0.263 & 28.14 & 32.31 & 23,809 & 0.216 & 23.12 & 25.43 \\
    RotatE & 3,593 & 0.211 & 23.32 & 35.86 & 3,992 & 0.265 & 33.51 & 43.23 \\
    \midrule
    ATransN & \bf 617 & \bf 0.280 & \bf 34.54 & \bf 49.22 & \bf 636 & \bf 0.321 & \bf 42.34 & \bf 55.96 \\
    \bottomrule
  \end{tabular}
\end{table*}

\subsection{Implementation}

The training process of ATransN is summarized in Algorithm~\ref{algorithm1}. 
The transition network $W$ that maps teacher entity embeddings into target entity embedding space consists of two linear layers to handle the complex transformation.
The generator $G$ that generates the noisy fake embeddings is a two-layer MLP with a LeakyReLU activation after the first layer; the discriminator $D$ that tries to distinguish whether two distributions are similar consists of one linear layer followed by a LeakyReLU activation and layer normalization, and one linear layer followed by a Sigmoid activation.
We select the best models based on the sum of $\frac{100}{\text{MR}}$, MRR, Hits@3, and Hits@10 on the validation data.
We use the same strategy for the teacher knowledge graph training.
Then we take the target entity and relation embeddings for the knowledge graph completion task.

Embeddings are initialized following the uniform distributions $\mathcal{U}(-\frac{\gamma+\epsilon}{n}, \frac{\gamma+\epsilon}{n})$ where $\epsilon$ is a hyperparameter fixed as 2~\cite{sun2019rotate}, $n$ is the dimension of the specific embedding module.
The dimension $n$ for TransE, DistMult, ComplEx, RotatE are set as 200, 200, 100, and 100 respectively for a fair comparison because the latter two are in the complex space, containing both real vectors and imaginary vectors.
Parameters of the transition network $W$ are initialized orthogonally~\cite{SaxeMG13}, while other parameters are initialized uniformly~\cite{HeZRS15}.

We choose TransE as the embedding module of ATransN and Adam~\cite{kingma2014adam} to optimize ATransN and other baselines.
The optimizer for the generator only updates parameters of $G$; the optimizer for the discriminator updates parameters of both $D$ and $W$; the optimizer for the embedding module updates parameters of knowledge graphs representations as well as $W$.
To mitigate the performance impact of adversarial and transfer learning on the target data, we add the cyclical cosine annealing scheduler for $\alpha$ and $\beta$~\cite{FuLLGCC19} and the learning rate scheduler to warm up the learning rates for the first 1\% steps.
Detailed hyper-parameter searching and settings are described in Appendix~\ref{appendix:hyper_parameters}.

\subsection{Discussion and Analysis}

\begin{table*}[!ht]
  \centering
  \caption{Model performance of ablation models. $\alpha=0$ means no distance constraint and $\beta=0$ means no triplet constraint.}
  \label{tab:ablation}
  \begin{tabular}{c|ccc|ccc|ccc|ccc}
    \toprule
    & \multicolumn{3}{c|}{CN3l (EN-DE)} & \multicolumn{3}{c|}{WK3l-15k (EN-FR)} & \multicolumn{3}{c|}{DWY100k (DBP-YG)} & \multicolumn{3}{c}{DWY100k (DBP-WD)} \\
    Model & MR & MRR & Hits@3 (\%) & MR & MRR & Hits@3 (\%) & MR & MRR & Hits@3 (\%) & MR & MRR & Hits@3 (\%) \\
    \midrule
    ATransN & \bf 446 & \bf 0.205 & \bf 33.76 & \bf 403 & 0.422 & \bf 45.75 & 617 & \bf 0.280 & \bf 34.54 & \bf 636 & \bf 0.321 & \bf 42.34 \\
    \quad $\alpha$ = 0 & 697 & 0.164 & 26.61 & 416 & 0.422 & 45.70 & 2,671 & 0.262 & 30.03 & 3,103 & 0.247 & 34.19 \\
    \quad $\beta$ = 0 & 470 & 0.202 & 33.35 & 408 & \bf 0.423 & \bf 45.75 & \bf 611 & 0.279 & 34.53 & \bf 636 & \bf 0.321 & \bf 42.34 \\
    \hline
    CTransE & 892 & 0.164 & 27.01 & 438 & 0.422 & 45.74 & 3,216 & 0.265 & 30.17 & 3,070 & 0.247 & 34.19 \\
    \quad $\alpha$ = 0 & 918 & 0.163 & 26.25 & 438 & 0.422 & 45.72 & 3,223 & 0.264 & 30.12 & 3,080 & 0.247 & 34.24 \\
    \quad $\beta$ = 0 & 915 & 0.163 & 26.80 & 439 & \bf 0.423 & 45.74 & 3,203 & 0.265 & 30.17 & 3,070 & 0.247 & 34.19 \\
    \hline
    TransE & 910 & 0.162 & 26.29 & 443 & 0.419 & 45.39 & 2,778 & 0.220 & 25.65 & 3,148 & 0.237 & 33.03 \\
    \bottomrule
  \end{tabular}
\end{table*}

\begin{figure*}[!ht]
    \centering
    \includegraphics[width=\linewidth]{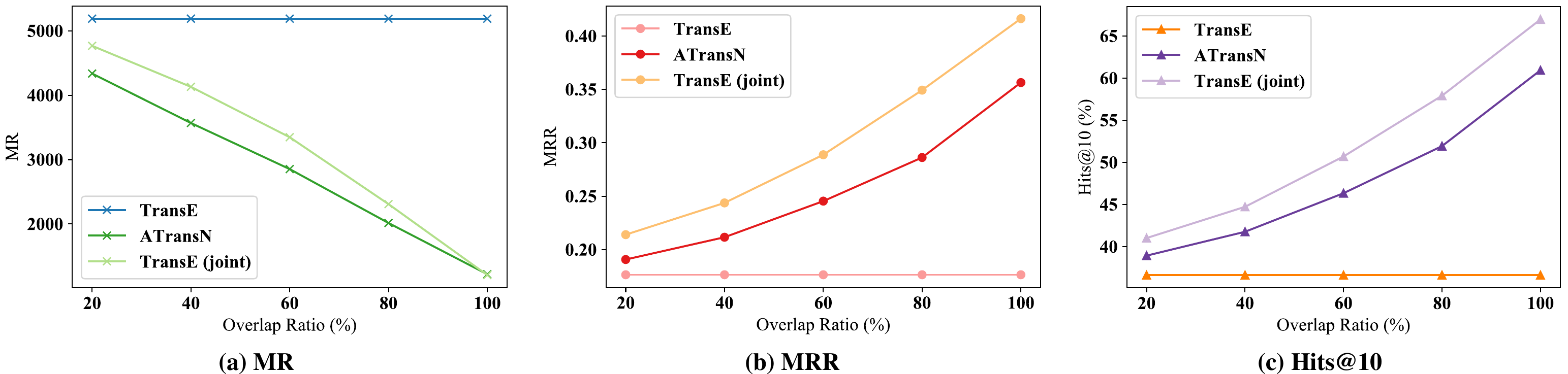}
    \caption{Performance improvement in different overlap ratios on subsets of DWY100k (DBP-WD).}
    \label{fig:ablation_ratio}
    \Description[Evaluation on MR, MRR and Hits@10]{(a) shows MR trends of TransE, ATransN and TransE(joint) as the entity overlap ratio increases, all of which decline in general. (b) and (c) describe MRR trends and Hits@10 trends of these three model settings seperately, all of which increase as the figures show.}
\end{figure*}

The empirical results on two standard multi-lingual datasets CN3l, WK3l-15k, and the multi-source dataset DWY100k are shown in Tables~\ref{tab:cn3l_wk3l} and~\ref{tab:dwy100k}. And teacher performance is listed in Appendix~\ref{appendix:teacher_performance}. These tables report MR, MRR, Hits@3, and Hits@10 results of four different baseline models and ATransN on each dataset.

We first compare our ATransN with the four baselines. ATransN has the best performance on WK3l-15k, DBP-YG, and DBP-WD across all metrics. On CN3l, ATransN outperforms all the baselines on all metrics except MRR. In this case, ATransN is still competitive with RotatE on MRR and significantly surpassing RotatE on other metrics. 
Comparing with TransE, ATransN improves the Hits@3 and Hits@10 by notable margins of 7.47\% and 10.72\% on CN3l, substantial margins of 8.89\% and 10.73\% on DBP-YG, and remarkable margins of 9.31\% and 11.80\% on DBP-WD.
In addition, ATransN can help make relevant head or tail entities come top in ranks, reflected by significant decreases of MR among all data.
This would be very helpful for the knowledge graph completion because models with transfer learning tend to have higher recall scores.
From these results, we show that ATransN successfully transfers knowledge from an auxiliary knowledge graph, and improves representation expressiveness on the target knowledge graph.

Second, the improvement of WK3l-15k is not as significant as others. There are two possible reasons. 
One is that compared with the other three target knowledge graphs, the French knowledge graph on WK3l-15k is much denser, where the average degree of entities is about 11 while others are no greater than 5. 
The representations learned on the target triplets have already been good enough. 
Thus, the auxiliary knowledge graph contributes little. 
The other is that the entity alignment ratio is quite small on the WK3l-15k data, while ratios on the other two datasets reach 90\% and even 100\%.
A smaller alignment ratio usually indicates a larger difference between the two knowledge graphs.

\subsection{Ablation Analysis}\label{sec:ablation}

We conduct extensive ablation studies to prove the effectiveness of each module in ATransN. 
To verify the validity of the distance constraint and the triplet constraint, models with the best $\alpha$ and $\beta$ are searched when the other hyper-parameter is set as 0.
To verify the necessity of the adversarial adaptation module, we implement a baseline \textbf{CTransE}, which only adds the two constraints in the training objective like Eq.~\eqref{eq:obj} with constant weights.
The consistency degree of two aligned entities is no longer considered in CTransE.

The experimental results on four different datasets are reported in the first block of Table~\ref{tab:ablation}. 
ATransN almost achieves the best performance when $\alpha > 0 $ and $\beta > 0$, which means two constraints in Eq.\eqref{eq:obj} really contribute to the embedding learning process. 
Furthermore, ATransN w/ $\beta=0$ has competitive results with ATransN on CN3l, WK3l, DBP-YG, and DBP-WD, which shows that the distance constraint plays a more important role in the transfer learning.
We find that ATransN w/ $\alpha=0$ can still outperforms TransE at the bottom.
The triplet constraint mainly improves MR on CN3l and WK3l but Hits@3 and Hits@10 on DWY100k.
The assumption behind the triplet constraint is too strong as it requires a teacher entity embedding to fit all triplets of its corresponding aligned entity in the target knowledge graph.
When data are not in the same domain, it does not work; when data are in the same domain, it may duplicate triplets.
On the contrary, the distance constraint seems looser as it only makes two aligned entity embeddings have similar directions instead of the same elements.
WK3l-15k is a denser and less-aligned dataset, and the results of ATransN w/ $\alpha=0$ are much closer to ATransN.
Two constraints perform similarly because target entity and relation representations can be trained well and supervisions from constraints are dispelled by target triplets.
In this case, the general contributions of the two constraints become roughly equal.

\begin{figure*}[!ht]
    \centering
    \includegraphics[width=0.96\linewidth]{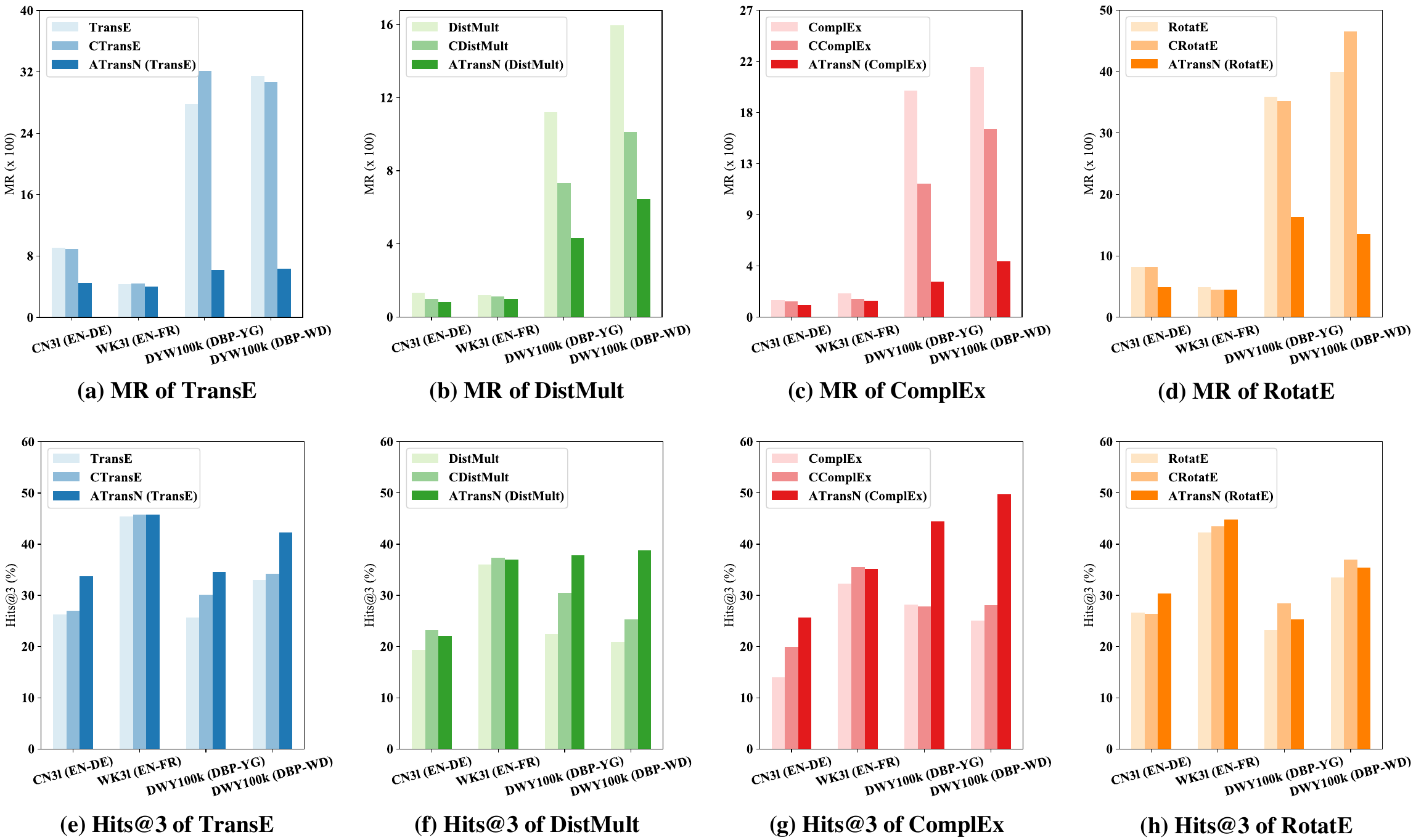}
    \caption{Performance improvement of four different KG embedding models.}
    \label{fig:extension_kge}
    \Description[MR and Hits@3 results with four different methods as embedding module on all datasets]{It includes 8 sub-figures for four different embedding modules, where (a,e) for TransE, (b,f) for DistMult, (c,g) for ComplEx, (d,h) for RotatE. (a-d) show the MR results and (e-h) show the Hits@3 results.}
\end{figure*}

When removing the adversarial adaptation module, numbers of CTransE in Table~\ref{tab:ablation} drop a lot on all metrics, which proves that measuring the consistency degree is vital during the transfer process.
In other words, the embedding transfer process without the adversarial adaption module is not very effective and even has a negative influence as the results of CTransE with $\alpha=0$ shown on CN3l.
Therefore, it is the predictions of the discriminator as the dynamic weights that bring significant improvement. 
For the CTransE baseline, we also report the best results when $\alpha$ and $\beta$ are zero respectively to explore how the distance constraint and the triplet constraint work during the transfer process.
On the two datasets in DWY100k, CTransE without the triplet constraint has the same results as CTransE.
On CN3l, the combination of two constraints is of help for the learning process. 
However, on the WK3l-15k dataset, combining the two constraints is not helpful.
Besides, CTransE w/o the triplet constraint performs better than CTransE w/o the distance constraint on most data, but the difference is small.
In conclusion, the distance constraint has a similar effect to the triplet constraint when the constraint weight is constant.
And the combination of the two constraints could affect each other in this case.

To further analyze how the entity overlap ratio would influence the experimental results, we design more experiments on subsets of DWY100k (DBP-WD). 
Figure~\ref{fig:ablation_ratio} shows performance curves on three metrics under 5 different overlap ratios \{20\%, 40\%, 60\%, 80\%, 100\%\}. 
The total number of entities in the teacher and target are 50,000 in each setting and we remain the same target knowledge graph for a fair comparison. 
The difference is the teacher part, where the aligned entities are transitive. That is, if an aligned entity pair appears in the aligned set of 20\% ratio, then it must exist in the set of 40\%, and so on.
As we can see, all metrics become better steadily as the overlap ratio increases.
Furthermore, to show the upper bound of improvement brought by introducing the teacher, we designed the \textbf{TransE(joint)}, which is trained on the target triplets and teacher triplets after id mapping.
Embedding module is learned on the merged training set, but the original target validation and test data are used to choose and report models.
The gaps between curves of ATransN and TransE(joint) corresponding to the same metric become smaller with the rise of the overlap ratio.
However, the margins between ATransN and TransE cannot be ignored gradually (in fact, curves of TransE are not changed because validation and test data are the same).
Thus, introducing the teacher knowledge graphs does improve target representation learning, and the performance grows rapidly as the entity overlap increases.

\begin{table*}[!ht]
  \centering
  \caption{Statistics of different teacher knowledge graphs for Wikipedia in English.}
  \label{tab:stts_wikipedia}
  \begin{tabular}{cc|ccccc}
    \toprule
    Teacher & Student & \#Entities & \#Relations & \#triplets & \#Aligned entities & Alignment ratio (\%) \\
    \midrule 
    DBpedia (WD) & & 100,000 & 330 & 463,294 & 12,555 & 12.56 \\
    & Wikipedia (EN) & 15,169 & 2,228 & 203,502 & 1,553 & 10.24 \\
    \hline
    Wikipedia (FR) & & 15,393 & 2,422 & 170,605 & 2458 & 15.97 \\
    & Wikipedia (EN) & 15,169 & 2,228 & 203,502 & 2,496 & 16.45 \\
    \hline
    Freebase & & 14,541 & 237 & 310,116 & 1,299 & 8.93 \\
    & Wikipedia (EN) & 15,169 & 2,228 & 203,502 & 3,320 & 21.89 \\
    \hline
    \textit{Multiple} & & - & - & - & - & - \\
    & Wikipedia (EN) & 15,169 & 2,228 & 203,502 & 4,155 & 27.39 \\
    \bottomrule
  \end{tabular}
\end{table*}

\begin{table}[!ht]
  \centering
  \caption{Model performance on different teacher knowledge graphs for Wikipedia in English. \textit{Multiple} means all three knowledge graphs are regarded as teachers.}
  \label{tab:wikipedia}
  \setlength{\tabcolsep}{3pt}
  \begin{tabular}{cc|cccc}
    \toprule
    Model & Teacher(s) & MR & MRR & Hits@3 (\%) & Hits@10 (\%) \\
    \midrule 
    TransE & - & 239 & 0.416 & 47.16 & 60.72 \\
    \hline
    \multirow{4}{*}{ATransN} & DBpedia (WD) & 238 & 0.418 & 47.38 & 60.93 \\
    & Wikipedia (FR) & 235 & 0.418 & 47.42 & 60.90 \\
    & Freebase & 238 & 0.418 & 47.48 & 60.94 \\
    & \textit{Multiple} & \bf 233 & \bf 0.420 & \bf 47.62 & \bf 61.22 \\
    \bottomrule
  \end{tabular}
\end{table}

\subsection{Extensions of ATransN}
\subsubsection{Combining Other Embedding Methods}
As the modules in ATransN are independent with each other and the transfer process is not related to specific methods, we can easily extend this framework with other knowledge embedding methods or adversarial networks.

In this section, we choose each of the remaining knowledge graph embedding models listed in Table~\ref{tab:baselines} as the embedding module, including DistMult, ComplEx, and RotatE. 
To distinguish different models, we mark the specific embedding method in brackets after ATransN. 
For example, ATransN with DistMult as the embedding module is denoted as ATransN (DistMult).
Moreover, we also conduct some ablation studies to further prove the necessity of the adversarial network.
We name the ablation setting the same way as CTransE.
For instance, ATransN (DistMult) w/o the adversarial adaptation module is denoted as CDistMult. 
We draw bar graphs according to specific evaluation results and report the results of the four knowledge graph embedding methods in Figure~\ref{fig:extension_kge}.

For the MR metric, Figure~\ref{fig:extension_kge} (a)-(d) show obvious trends, of which the lower value means the better performance.
As we can see, ATransN with different embedding methods in darker colors always achieves the best performance on the four datasets.
Moreover, CDistMult and CComplEx decrease the MR further.
The most likely reason is two constraints are appropriate to such bilinear score functions because all of them involve element-wise product among entity embeddings.
However, the Hadamard product in complex space of CRotatE is not fully aligned with the cosine distance.
To be more accurate, the cosine distance adds all element-wise product but the Hadamard product have both additions and subtractions.

For the Hits@3 metric plotted in Figure~\ref{fig:extension_kge} (e)-(h), ATransN almost outperform all baselines significantly, proving that adversarial transfer learning does work.
Most of them also outperform ATransN w/o adversarial adaptation modules.
The performance degradation of ATransN (DistMult) and ATransN (ComplEx) is very little on WK3l. 
But both of them boost significantly on DWY100k.
Hence, the two models are good at larger data with fewer relation types.
However, ATransN (RotatE) and CRotatE are still not consistent.
How to extend adversarial transfer learning to complex spaces is still worthy of exploring.

\subsubsection{Multiple Teacher Transferring}

To further explore the extensibility of our ATransN, we conduct another experiment where there are three different teacher knowledge graphs for embedding transfer. 
To be specific, we make use of DBpedia (WD), Wikipedia (FR) and FB15k-237~\cite{toutanova2015observed} as three teachers and learn knowledge graph representations for Wikipedia (EN). 
In this challenging setting, teacher knowledge graphs include both a knowledge graph in a different language and two knowledge graphs from different data sources, and the target knowledge graph is pretty dense.
We separately train a transition network and an adversarial adaptation module for each teacher knowledge graph.
These modules are combined with one embedding module to learn the target knowledge graph representations.
The statistics of these knowledge graphs are summarized in Table~\ref{tab:stts_wikipedia}.
As we can see, the target knowledge graph Wikipedia (EN) is much denser than DBpedia (WD) and Wikipedia (FR), while is sparser than Freebase.

Table~\ref{tab:wikipedia} shows the experimental results on Wikipedia in English. 
The performance of TransE and ATransN models with a single teacher is shown in the first four rows.
No matter what the teacher is, ATransN can make a slight improvement.
As the alignment ratio increases, Hits@3 and Hits@10 receive further gains.
And we could conclude that ATransN can be applied to both knowledge graphs in different languages and those from different sources.
Besides, it is not required that the teacher knowledge graph is denser than the target knowledge graph.
Both make ATransN be able to apply to many scenarios.
Furthermore, combing three different teacher knowledge graphs obtains a higher ratio of the entity alignment and better evaluation results on the knowledge graph completion task.
This means the transfer learning process from different teacher knowledge graphs would not influence each other.
So we can collect entity embeddings from various teacher knowledge graphs to further improve the knowledge representation learning in practice.

\subsection{Hyper-parameter Sensitivity}

\begin{figure*}[t]
    \centering
    \includegraphics[width=0.95\textwidth]{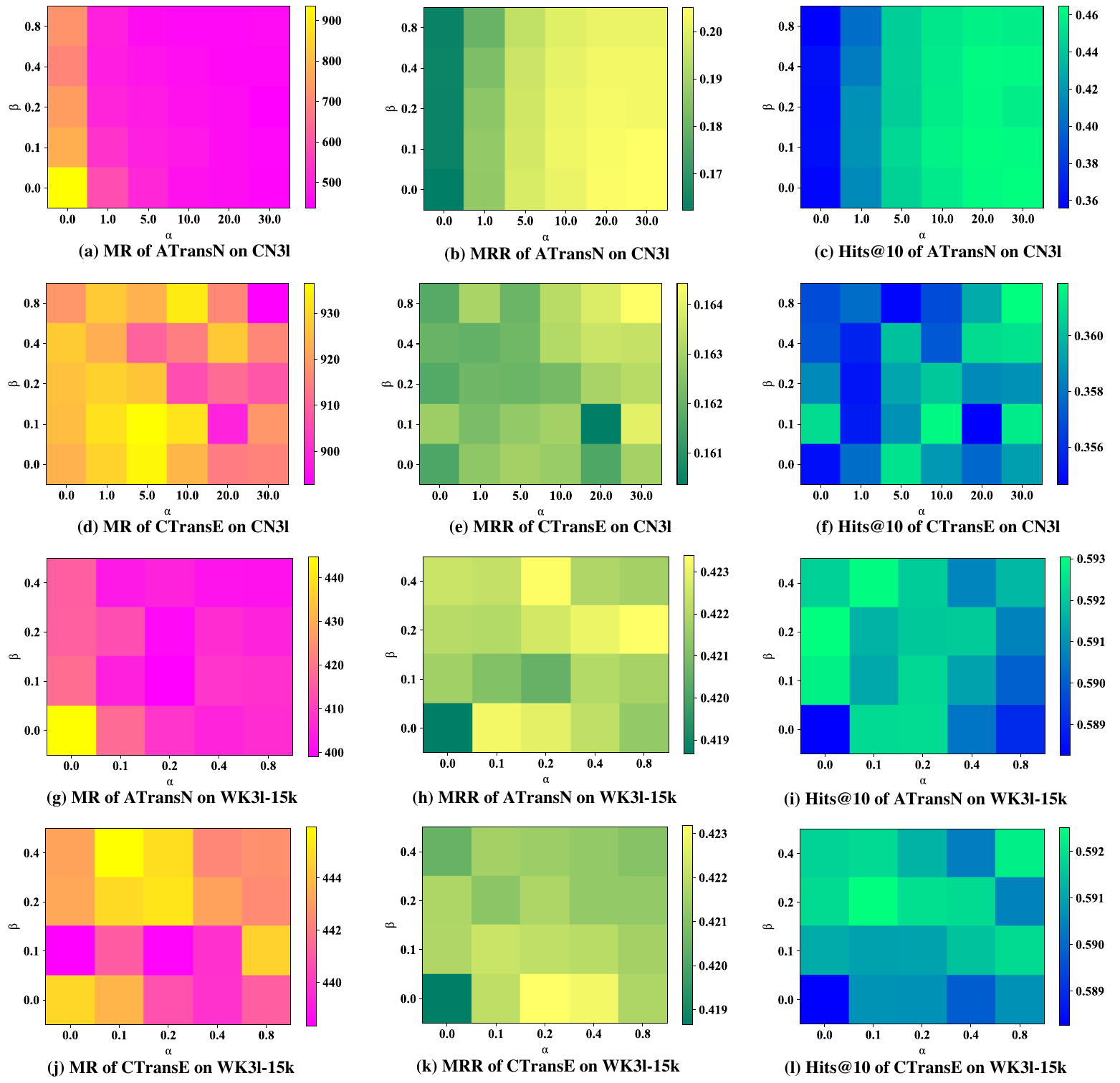}
    \caption{Hyper-parameter sensitivity on CN3l (EN-DE) and WK3l-15k (EN-FR). (a)-(f) are results of CN3l (EN-DE), while (g)-(l) are results of WK3l-15k (EN-FR); (a)-(c) and (g)-(i) correspond to ATransN, while (d)-(f) and (j)-(l) correspond to CTransE.}
    \label{fig:grad}
    \Description[Heap maps of hyper-parameters for the constraints on CN3l (EN-DE) and WK3l-15k (EN-FR).]{See the caption.}
\end{figure*}

Two important hyper-parameters $\alpha$ and $\beta$ are involved for constants in our previous experiments. 
It is unknown how sensitive the performance of ATransN is to these two parameters. 
Thus, we perform sensitivity analysis in this section to explore the effect of $\alpha$ and $\beta$ in our framework.
We choose ATransN and CTransE and two representative datasets CN3l and WK3l-15k.
Figure~\ref{fig:grad} provides heat maps of results on three metrics, namely, MR, MRR, and Hits@10.

On CN3l, we consider the weight of the distance constraint $\alpha \in \{0, 1, 5, 10, 20, 30\}$, the weight of the triplet constraint $\beta \in \{0.0, 0.1, 0.2, 0.4, 0.8.\}$.
It is easy for us to see the trend that a larger $\alpha$ can result in better performance.
When $\beta$ increases, MR becomes better but MRR becomes worse. It is also clear to see the best $\beta$ value for Hits@10 is around 0.1 or 0.4.
This proves that the distance constraint is effective enough to transfer entity features but the triplet constraint can slightly adjust.
However, from the results of CTransE on CN3l, the most immediate observation is each metric has its own optimal configuration.
And it is infeasible to find a pattern for Hits@10.
That suggests that $\alpha$ and $\beta$ become very sensitive without the adversarial adaptation module. 
It requires careful hyper-parameter searching.
But CTransE with even the best configuration is still far worse than ATransN with a proper but not the best configuration.

As the entity alignment ratio on WK3l-15k is much smaller, we consider $\alpha \in \{0.0, 0.1, 0.2, 0.4, 0.8\}$ and $\beta \in \{0.0, 0.1, 0.2, 0.4\}$. 
Heat maps plotted in the third line of Figure~\ref{fig:grad} is not as regular as the first line on CN3l, but we can still find some trends on MR and MRR. 
In general, the best value of $\alpha$ is between 0.1 and 0.2, while the best value for $\beta$ is about 0.4.
Besides, there is no consistent pattern in heat maps of CTransE in the last line, let alone the performance.
In conclusion, the two parameters of ATransN is not as sensitive as those of CTransE.
And the adversarial adaptation module makes ATransN appropriate to different scenarios in a general way.

\section{Conclusion}
We propose an adversarial transfer network (ATransN) and demonstrate its effectiveness in the context of the knowledge graph completion task. 
ATransN successfully transfers features in the teacher knowledge graphs to target ones on three different datasets.
Extensive ablation studies prove the effectiveness and necessity of modules in ATransN, including different constraints in the embedding transfer module and the dynamic consistency score in the adversarial adaptation module.
At the same time, ATransN is also a general framework that can extend to other shallow embedding models and multiple teacher knowledge graphs.
It would be worthwhile to explore entity alignment techniques in the future.

\begin{acks}
This work was supported by the Special Funds for Central Government Guiding Development of Local Science \& Technology (No. 2020B1515310019) and the National Natural Science Foundation of China (U1711262, U1711261 and No. 62006083).
\end{acks}

\clearpage

\bibliographystyle{ACM-Reference-Format}
\bibliography{atransn.bib}

\clearpage

\appendix

\section{Hyper-parameters}
\label{appendix:hyper_parameters}

We train ATransN as well as baselines in mini-batches for at most 300 epochs. 
We manually set the batch size $N_l$ to make one training epoch finished in approximately 100 steps. Thus, the batch sizes for CN3l, WK3l-15k, and FB15k-237 are 128, 1024, and 1024 respectively. However, we cannot set larger values for DWY100k due to the GPU memory limitation so that the batch size for DWY100k is also 1024.

For the embedding module, we select the learning rate $lr_e$ among $\{1e-2, 5e-3, 2e-3, 1e-3, 5e-4, 2e-4, 1e-4\}$ and the margin $\gamma$ among $\{1.0, 2.0, 4.0, 8.0, 16.0, 32.0\}$;
for the adversarial adaption module, we search the learning rate $lr_a$ among \{5e-5, 1e-4, 2e-4, 5e-4, 1e-3\}.
$\beta$ is chosen among $\{0.0, 0.1, \cdots, 1.0\}$ for all datasets because it does not make sense when the framework pays more attention on the transferred triplets instead of the target triplets in the knowledge graph itself.
But considering the density of the target knowledge graph as well as the alignment ratio of entities, we seek the best $\alpha$ among $\{0.0, 1.0, 5.0, 10.0, 20.0, 30.0\}$ for CN3l, $\{0.0, 0.1, 0.2, 0.4, 0.8\}$ for WK3l-15k and FB15k-237, $\{0.0, 1.0, 5.0, 10.0\}$ for DWY100k.
We fix generator training steps $T_g$ as 5, discriminator training steps $T_d$ as 5, minibatch size for adversarial modules $N_a$ as 128 for all datasets.

When searching the negative sampling size $k$, we find that a larger sampling size usually results in better performance but requires more time to converge. This hyper-parameter is fixed as 128 for all datasets.
Models are the most robust when $lr_e$=1e-3 and $lr_a$=2e-4.
We conduct experiments of TransE models to find the best $\gamma$ values and then search $\alpha$ and $\beta$ for ATransN. For Distmult and ComplEx, $\gamma$ is set as one because $\gamma$ does not have a significant effect on the final performance of the two models.
We also search hyper-parameters $\gamma$ for RotatE and find a larger $\gamma$ usually results in better performance.
For the CN3l dataset, the optimal hyper-parameters of ATransN are $\gamma$=8.0, $\alpha$=30, $\beta$=0.1;
for the WK3l-15k dataset, the optimal hyper-parameters of ATransN are $\gamma$=4.0, $\alpha$=0.1, $\beta$=0.4.
The optimal configuration of ATransN for DWY100k datasets are $\gamma$=16.0, $\alpha$=5.0, $\beta$=0.1, and that for the FB15k-237 dataset is $\gamma$=8.0, $\alpha$=0.1, $\beta$=0.1.

\pagebreak

\section{Teacher Performance}
\label{appendix:teacher_performance}

\begin{table}[!h]
  \centering
  \caption{Teacher performance of different embedding models.}
  \begin{tabular}{c|cccc}
    \multicolumn{5}{c}{Subtable (a): CN3l} \\
    \toprule
    Model & MR & MRR & Hits@3 (\%) & Hits@10 (\%) \\
    \midrule
    TransE & 451 & 0.214 & 29.93 & \bf 42.24 \\
    DistMult & 665 & 0.238 & 28.12 & 38.79 \\
    ComplEx & 810 & 0.241 & 29.62 & 37.74 \\
    RotatE & \bf 423 & \bf 0.240 & \bf 31.27 & 42.20 \\
    \bottomrule
    \multicolumn{5}{c}{} \\
    \multicolumn{5}{c}{Subtable (b): WK3l-15k} \\
    \toprule
    Model & MR & MRR & Hits@3 (\%) & Hits@10 (\%) \\
    \midrule
    TransE & \bf 239 & 0.416 & \bf 47.16 & \bf 60.72 \\
    DistMult & 611 & \bf 0.418 & 46.53 & 57.45 \\
    ComplEx & 1,389 & 0.372 & 40.26 & 48.22 \\
    RotatE & 296 & 0.411 & 46.90 & 59.80 \\
    \bottomrule
    \multicolumn{5}{c}{} \\
    \multicolumn{5}{c}{Subtable (c): DWY100k (DBP-YG)} \\
    \toprule
    Model & MR & MRR & Hits@3 (\%) & Hits@10 (\%) \\
    \midrule
    TransE & \bf 2,957 & \bf 0.203 & \bf 26.26 & \bf 39.50 \\
    DistMult & 13,255 & 0.151 & 16.43 & 23.88 \\
    ComplEx & 22,787 & 0.141 & 15.17 & 20.03 \\
    RotatE & 4,152 & 0.186 & 22.14 & 35.36 \\
    \bottomrule
    \multicolumn{5}{c}{} \\
    \multicolumn{5}{c}{Subtable (d): DWY100k (DBP-WD)} \\
    \toprule
    Model & MR & MRR & Hits@3 (\%) & Hits@10 (\%) \\
    \midrule
    TransE & \bf 2,567 & 0.272 & 35.18 & \bf 47.55 \\
    DistMult & 15,849 & 0.181 & 20.61 & 28.62 \\
    ComplEx & 20,611 & 0.239 & 25.66 & 29.83 \\
    RotatE & 4,160 & \bf 0.303 & \bf 36.10 & 46.69 \\
    \bottomrule
  \end{tabular}
\end{table}

\end{document}